\documentclass[11pt,a4paper]{article}
\usepackage[hyperref]{acl2020}
\usepackage[utf8]{inputenc}
\usepackage{times}
\usepackage{latexsym}

\usepackage{graphicx}% http://ctan.org/pkg/graphicx

\usepackage{microtype}

\aclfinalcopy % Uncomment this line for the final submission
\def\aclpaperid{***} %  Enter the acl Paper ID here
\title{Processing M.A. Castrén's Materials:\\ Multilingual Typed  and Handwritten Manuscripts}

\author{Niko Partanen\textsuperscript{[0000-0001-8584-3880]} \\
  Department of Finnish, \\Finno-Ugrian and Scandinavian Studies \\
  University of Helsinki \\\And
  Jack Rueter\textsuperscript{[0000-0002-3076-7929]} \\
  Department of Digital Humanities \\
  University of Helsinki \\\AND
  Mika Hämäläinen\textsuperscript{[0000-0001-9315-1278]} \\
  Department of Digital Humanities \\
  University of Helsinki \& Rootroo Ltd \\\And
  Khalid Alnajjar\textsuperscript{[0000-0002-7986-2994]} \\
  Department of Digital Humanities \\
  University of Helsinki \& Rootroo Ltd \\\AND
  \normalfont{\texttt{firstname.surname@helsinki.fi}}}

\date{}

\begin{document}
\maketitle
\begin{abstract}

The study forms a technical report of various tasks that have been performed on the materials collected and published by Finnish ethnographer and linguist, Matthias Alexander Castrén (1813–1852). The Finno-Ugrian Society is publishing Castrén's manuscripts as new critical and digital editions, and at the same time different research groups have also paid attention to these materials. We discuss the workflows and technical infrastructure used, and consider how datasets that benefit different computational tasks could be created to further improve the usability of these materials, and also to aid the further processing of similar archived collections. We specifically focus on the parts of the collections that are processed in a way that improves their usability in more technical applications, complementing the earlier work on the cultural and linguistic aspects of these materials. Most of these datasets are openly available in Zenodo. The study points to specific areas where further research is needed, and provides benchmarks for text recognition tasks. % some of the tasks that have been conducted. %, especially in relation to text recognition. 

\end{abstract}

\section{Introduction}

As a research domain, the Natural Language Processing has regularly focused on the formal written varieties of the most widely used languages of the world. At the same time there has been a growing interest in both non-standard and informal language \cite{4c94883ec360438dbd36b59880b06331,537e4ad5fc054610a23474d91532bb07}, and their historical varieties \cite{f584ffd3bf9d4289b77d90d549089df6,698e8319ca394a758cf590e054945a35}. 
%jaska: seuraava lause on vaikea ymmärtää
% Niko: Joo pitää kehittää
The research potential of historical language varieties is clearly on the upbound, and one can argue that the need is already quite evident, as digitization processes in libraries and archives around the world have reached relatively mature stages and already have large digital collections available. %The work described in this study has also partially taken place in a context where a scientific society has started to systematically publish and process some of the resources in their collections. 
%Such materials are also becoming increasingly available in digital format, as digitization processes of libraries and archives have in many cases reached relatively late stages with large digital collections already available. The work described in this study has also partially taken place in a context where a scientific society has started to systematically publish and process some of the resources in their collections. 

Finnish ethnographer and linguist Matthias Alexander Castrén (1813–1852) produced a large collection of field notes, and also published widely on languages of Northern Eurasia. Recently, two hundred years had passed since his birth, and in this connection the Finno-Ugrian Society launched a project where several of his field notes and grammars are published as commented editions, available both digitally and in print. Numerous monographs have already been published in the series \cite{c1,c2,c3,c4,c5,c6,mcepi}. The complete series will contain more than twenty volumes. %, and we currently envision that digital methods will become more strongly adapted in the researchers' workflows while the project develops. 
This article discusses the processing of original raw materials up to this point, with a goal of setting a vision of how this process can be refined later on. 

Within the research tradition of the Uralic languages, Matthias Castrén is often renowned as the most significant Finnish linguist of the 19th century. Castrén collected vast materials from almost thirty languages on his expeditions to Lapland and Northern Russia
%jaska: this is problematic in English that we are naming an entity and then one of its subentities, isn't it?
% Niko: Onks parempi?
%jaska: ... on his expeditions to Lapland and Northern Russia, Olen vain vaihtanut järjestystä ja kaventanut Venäjää
between 1838 and 1849 \cite[15]{mcpreface}. % This cannot really refer to Castrén himself?
The materials are stored in the National Library of Finland. The number of handwritten manuscript pages is approximately ten thousand. %For some languages, such as Yeniseic Kott, Castrén's field notes are among the only existing materials that were documented before the language ceased to be spoken. 
Castrén's work carries a unique historical dimension for the languages he studied, and his manuscripts and extensive correspondence with other researchers of the time are also valuable for the history of scientific research. 

Our study presents individual datasets built from Castrén's materials and reports benchmarks on various text recognition experiments. The main repository for related data is Manuscripta Castreniana collection in Zenodo\footnote{\url{https://zenodo.org/communities/castreniana}}, and other locations are specified when datasets are discussed. We also discuss individual experiments with the text recognition of Castrén's unpublished and published materials, and contextualize the results more widely within early linguistic descriptions. We analyse some of the challenges met in further processing of this content, and delineate possible ways forward. The most important step we can identify is making these materials better available, so that further work can build upon the contributions of more researchers. This is also the step we are trying to help make. We would hope, for example, that eventually Castren's materials would be included in different shared tasks. In the same spirit we also share all our processing code in GitHub\footnote{\url{https://github.com/nikopartanen/manuscripta}}, which we hope makes these materials easier to access for different researchers in the digital humanities and related fields.

\section{Related work}

Historical dataset creation is one topic that connects closely to ours. Especially within the Universal Dependencies project \cite{11234/1-3687} there are numerous instances of historical language treebanks. There are five Latin treebanks, Old Church Slavonic treebank \cite{haug2008creating}, Old Turkish \cite{ud_old_turkish_tonqq_2020} and Old French treebank \cite{stein:halshs-01122079}, just to mention some of them. Such resources are in a central role, as they allow training NLP models to address different downstream tasks for these language varieties. 
Naturally, any openly available resource, in plaintext or with annotations, can be used for these purposes. 
At the same time, the CoNLL-U file format offers a good and well understood structure that can easily be compared.  %In this context we can mention… \citet{keersmaekers2020creating}… 

There are examples of such datasets being used in downstream tasks, such as lemmatizers and POS taggers created for Latin \cite{thibault_clerice_2021_4661034} and Old French \cite{camps2021corpus}. Work has been done also on Old Swedish (for example, \cite{borin2008something,adesam2016old}, but an actual diachronic corpus seems to be still under construction \cite{petterssontowards}. 
If such resources existed, the analysis of Castrén's 19th century Swedish would be in a different state. 
There is one unannotated diachronic corpus of Old Literary Finnish \cite{VKSkielipankki} and one morpho-syntactically annotated corpus of Mikael Agricola's works \cite{agricola-v1-1-korp_fi}. 
The latter has already been used to develop a lemmatizer as well \cite{agricolalemmatizer}. 
Named entity recognition (NER) for historical publications in Finnish has also received attention lately \cite{kettunen2017names,db30d8808461479b8944996a2e44778d}. 
A recent survey by \citet{humbel2021named} reviewed different named entity recognition systems for early modern textual documents. %, which is at least partially the document type represented here too. 
Their conclusion was that benchmarking different NER systems in this domain is not currently possible, and suggest wider use of shared forums such as computational linguistics conferences as one way to coordinate further discussion and practices. 
Study by \citet{idziak2021scalable} where Polish lexicographic cards were recognized and organized is in some aspects also close to what we would hope to achieve with materials discussed here. 
To our knowledge, there are no datasets, NLP tools or resources of historical varieties of the endangered languages included in these collections, especially in Castrén's writing system that is essentially an inauguration of a Latin based transcription (Latin transcription with some Cyrillic characters). 

% Third, we also must recognize that scholarly work and research into the content of these documents and manuscripts is very important for this discussion. Applying NLP methods into these materials improves both usability and findability, but if there is no use or research tradition, then it is unclear what is the motive of such work. Naturally, it is also possible that the further work emerges only after the resources are made more accessible. However, we want to point out some of the recent research done on Castrén's fieldnotes…

\section{Materials}

We discuss four sections of Castrén's materials. The first consists of ethnographic field notes in 19th century Swedish under the title \textit{Ethnographiska, historiska och statistiska anmärkningar}. Castrén wrote these texts in a extensive area that belongs to the northern regions of the contemporary Russian Federation. 
%jaska: Nenetsia? the Nenets Realm? sitten jätetään sellaisena: Nenetsia & Cyberstan :-)
% EEEEIIII laitetaan jotain järkevää
%with different topics ranging from Nenetsia to Siberia. 
This text is also multilingual, with numerous expressions in Cyrillic, but we can approach it largely as a Swedish text. This subset contains 188 pages of handwritten texts. We use this dataset in text recognition experiments reported below, but these materials will be added to the Zenodo collections at a later stage. 

\begin{figure*}[!ht]
\centering
  \includegraphics[width=10cm]{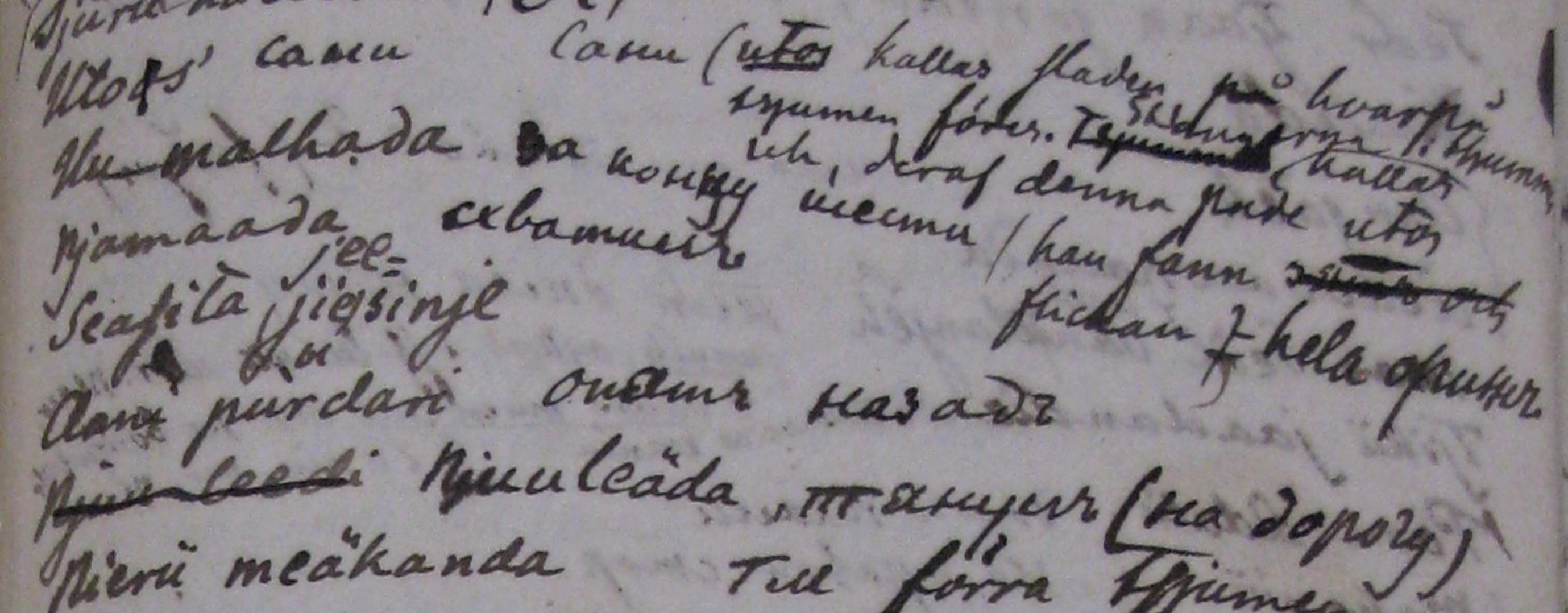}
  \caption{\href{https://www.sgr.fi/manuscripta/items/show/453}{Manuscripta Castreniana, Epic poem 1A, Page 155}}
  \label{castrensample}
\end{figure*}

The second dataset comes from Tundra Nenets epic poems that have a Russian translation with Swedish commentary. The Figure~\ref{castrensample} displays the typical structure in this manuscript. The page is split into two loosely distinguished columns, with Tundra Nenets transcription on the left and the Russian translation on the right. In the upper right region we see a comment in Swedish in parentheses, but there are also parenthetical clarifications in Russian, as seen in the bottom right corner. All in all, the material comprises 192 pages. This example also provides a good illustration of how the layout detection of these manuscripts is an additional challenge. This dataset is published as is in Zenodo \cite{castren_m_a_2021_5759599}. % 
The texts have been aligned line by line into the microfilm scans of the original manuscripts in collaboration between the University of Innsbruck and the Finno-Ugrian Society, and this material is an excellent test set for various tasks including text to image alignation, line segmentation and handwritten text recognition.  

The third dataset contains published Komi-Zyrian grammar of Castrén \cite{castren1844elementa} that is written in Latin. The grammar is 174 pages of printed text, all together. In the Manuscripta Castreniana project an English translation with commentary will be published, which adds a new dimension to what kind of computational tasks could be studied with this collection. 
Additionally, this partially proofread dataset is located in Zenodo \cite{niko_partanen_2021_5761573}, with 26 proofread pages of which 3 contain manually constructed tables. 
Thereby, this dataset is an example of 19th century printed Latin linguistic description, but also serves as the ground truth data for table layout detection as several tables are included with defined table cell structure. 
This grammar is also available as two different scans, both archived in Zenodo. 

The fourth dataset contains Castrén's Komi-Zyrian wedding laments and their transliteration in the modern Komi orthography \cite{niko_partanen_2021_5761135}. These materials were published with Finnish and German translations by \citet{aminoff1880a}, and our dataset contains aligned versions of the translations and different transcriptions. Similar dataset could also be created from Castrén's translation of the Gospel of St. Matthew. Crucially, Castrén's transcription system cannot be automatically converted into current orthography as it does not contain all phonemic information that the orthography does. However, the dataset, in itself, is very illustrative of a wider problem in applying NLP to these kinds of materials: the textual representation used has such a different level that, if we cannot transform the transcription into a more modern writing system, we cannot access the text with any current tools. 

\section{Text recognition}

\subsection{Background}

Text recognition of historical handwritten documents has advanced rapidly in the past few years. The Transkribus platform \cite{kahle2017transkribus} is leading the field in usability and adoption, and there are reports of consistent results. These include materials by authors such as Foucalt \cite{massot2019transcribing}, Eugène Wilhelm \cite{schlagdenhauffen2020optical}, Jeremy Bentham \cite[959]{muehlberger2019transforming} and Konstantin Rychkov \cite{arkhipovusing}. %This platform was used also by the Finno-Ugrian Society to process Castrén's text, and in the light of previous work it is not surprising that also our results were successful. 
%There are some aspects that possibly make Castrén's field notes differ, however, from the materials mentioned above. One is the multilinguality. 
As mentioned, Castrén's texts include dozens of languages, and Russian, Swedish and Latin are all used as meta languages in different contexts. 
Similarly presence of different writing systems is also a feature, and challenge, of datasets mentioned above, both with mixed Latin and Greek characters \cite[4]{schlagdenhauffen2020optical} and Evenki–Russian mixed content \cite{arkhipovusing}. 
However, the wide array of endangered languages is still a very specific feature of Castrén's materials. 
% However, the fact that Transkribus was successfully used in those earlier cases makes it unsurprising that our results were positive, too. 

Currently all text recognition experiments with Castrén's data have been done using the Transkribus platform. The reason for this has been that it allows collaborative editing, and has, at least for handwritten materials, been the currently leading platform. In our further processing the data from Transkribus is exported in Page XML format, which in our experience has been very satisfactory. %We'll discuss below some layout related issues. 
It appears that Castrén's materials are still particularly challenging to process, and we aim to delineate some of the more technical reasons next. 

%For text recognition Castrén's handwriting appears to be in the more challenging spectrum of such materials. Even in our best experiments the character accuracy has not been above 90\%, which is relatively low when compared to similar studies. We believe the reasons for low accuracy are partly related to the workflow adapted, and we introduce here some ways to possible improve the results. 

The first part of the materials was aligned with the microfilm images from XML files where one unaligned transcription version already existed. As these transcriptions were done outside Transkribus, with no visual connection to the actual documents, there may be features in the transcription that should be revised. At the same time the transcriptions were done before the text recognition task was even possible, so the character choices were primarily based on what was convenient for the individual researchers. When tens of thousands of pages are analysed together, it would be important to give careful consideration to which characters should be used to represent which of Castrén's special characters. This work is partially technical and a matter of deciding the correct Unicode characters, but also relates to linguistic analysis. The analysis of the latter type was also conducted for Evenki by \citet{arkhipov2021a}. % and steps should be taken towards harmonization. 

As these early versions have been aligned with microfilm scans, and only later have the better quality versions been scanned from the original documents, it may become necessary to realign the transcriptions with these more accurate versions. %At the same it is unclear whether the text recognition in itself is sensitive to such image quality and source differences, but we must still recognize the different images as one source of disparity in the dataset. 
%Even in their current state the documents stored in Zenodo are perfectly usable and fit for various tests and measurements along these lines. 
The materials have been arranged so that such a task is in principle feasible. 
The second dataset discussed in this study contains exactly these aligned microfilm scans, which, we believe could be used to measure both the impact of chosen character conventions and the quality of scans to the recognition result. 
The higher quality images are also stored in Zenodo with extensive metadata about the page content \cite{castren_m_a_2021_4782909}. 
Generally it is very typical for Castrén's materials that the same text exists in multiple versions. 
It is unclear to the current authors how to best connect these versions, but we see potentially high value in such an undertaking. 

\subsection{Experiments}

%We report here that the Finno-Ugrian Society's best Castrén text recognition model reaches currently accuracy of 93.3\%. 
In the current workflow, all texts are manually verified. The ground truth material increases continuously, and has now reached 358 pages. This includes 19,490 lines and approximately 57,000 words. The text recognition accuracy has not significantly improved when the last hundred pages have been added, and the accuracy has been hard to improve further. %In the recent experiments with Castrén's Saami materials we additionally see a drop of accuracy as there are characters in these materials not used elsewhere. 
We first discuss the results with Castrén's printed materials shown in Table~\ref{tab:ocr} and then discuss the handwritten text recognition results shown in Table~\ref{tab:htr}. 

Castrén's Komi-Zyrian grammar is written in Latin and it contains individual Komi words and experssions plus some comments in Russian. As Transkribus already contains numerous text recognition models for printed texts, the ideal scenario would be to use some of these directly. We compared some of the Transkribus models for printed texts against the proofread materials, the result being presented in Table~\ref{tab:ocr}.

\begin{table}[]
\centering
\small
\begin{tabular}{lll}
Model                                       & CER \%                          & WER \%                          \\ \hline
\multicolumn{1}{|l|}{Transkribus print 0.3} & \multicolumn{1}{l|}{0.91}  & \multicolumn{1}{l|}{4.60}  \\ \hline
\multicolumn{1}{|l|}{Noscemus GM 5}         & \multicolumn{1}{l|}{1.68}  & \multicolumn{1}{l|}{8.05}  \\ \hline
\multicolumn{1}{|l|}{German Kurrent 17th-18th} & \multicolumn{1}{l|}{9.26} & \multicolumn{1}{l|}{38.70} \\ \hline
\multicolumn{1}{|l|}{Acta 17 (extended)}    & \multicolumn{1}{l|}{10.10} & \multicolumn{1}{l|}{40.23} \\ \hline
\end{tabular}
\caption{\label{tab:ocr}Accuracy on printed Komi-Zyrian grammar written in Latin.}
\end{table}

Although Transkribus print 0.3 model\footnote{https://readcoop.eu/model/transkribus-print-multi-language-dutch-german-english-finnish-french-swedish-etc/} does not even include Latin, it still performs extremely well in our test scenario. In the model's documentation CER of 1.6\% is reported, and in our experiment the result was even better than that. This has wide significance for work on printed Latin texts, as the out-of-the-box tool truly gives functional result. This should be taken into account when planning further work on printed materials. 
As expected, the Russian words did not get recognized, and the printed model could benefit from wider inclusion of scripts. 
%jaska: eikö tämä ole melko tavallinen ongelma, että puuttuu vapaita kyrillisia fontteja. Oli sellainen ongelma kun Miikka kokeili tesseractia ja fst:tä aikoinaan ersäläisiin teksteihin.

With the handwritten materials the situation is different. We can see in the Table~\ref{tab:htr} that none of the available HTR models for the Swedish language work very well, even though the result on the Count Records model from the National Archives of Finland is relatively good. As this model is contemporary with Castrén, and also contains handwritten Swedish, the accuracy is not necessarily surprising. Yet, it tells that even with a handwritten text recognition model we do not need to start entirely from scratch. 

\begin{table}[!h]
\centering
\small
\begin{tabular}{lll}
Model                                         & CER \%                         & WER \%                          \\ \hline
\multicolumn{1}{|l|}{Castrén (+ NAF base model)}            & \multicolumn{1}{l|}{13.19} & \multicolumn{1}{l|}{35.01}       \\ \hline
\multicolumn{1}{|l|}{Castrén (no base model)}            & \multicolumn{1}{l|}{15.40}       & \multicolumn{1}{l|}{40.90}       \\ \hline
\multicolumn{1}{|l|}{NAF Court Records M10}   & \multicolumn{1}{l|}{28.65} & \multicolumn{1}{l|}{54.34} \\ \hline
\multicolumn{1}{|l|}{Gothenburg Police Reports} & \multicolumn{1}{l|}{32.09} & \multicolumn{1}{l|}{60.50} \\ \hline
\multicolumn{1}{|l|}{Edelfelt M13+}           & \multicolumn{1}{l|}{34.66} & \multicolumn{1}{l|}{63.59} \\ \hline
\multicolumn{1}{|l|}{Stockholm Notaries}        & \multicolumn{1}{l|}{43.82} & \multicolumn{1}{l|}{81.23} \\ \hline
\multicolumn{1}{|l|}{Jaemtlands domsagas M1+} & \multicolumn{1}{l|}{44.34} & \multicolumn{1}{l|}{78.43} \\ \hline
\end{tabular}
\caption{\label{tab:htr}Accuracy on Castrén's handwritten Swedish}
\end{table}

% The comparison with different Swedish models motivated an experiment where the best model, NAF Court Records M10 was used as a base model. 

Even if we were to try to use other models as base models in training, the gains would be relatively minor. Training the Castrén's HTR model with Court Records M10 from the National Archives of Finland as a base model does improve the CER by some percentages compared to Castrén's Ground Truth alone, and on the WER level the difference is almost five percentage points. We are not seeing entirely transformative differences in the results, but still there is a significant improvement that we get essentially for free. 

\section{Processing tools}

We have archived our processing scripts on GitHub and Zenodo so that they would be maximally useful for a wider community of researchers. Text recognized materials from Transkribus can be exported in Page XML format. The structure is highly standardized, but also relatively complicated. %There are some core tasks with text and line image extraction for which we share our processing code. %We provide essential scripts to parse the text content and, which are strongly connected to extracting the textual information at different document levels or extracting the corresponding image regions by pixel coordinates. 
We provide methods to read the lines and their bounding boxes from the XML files into a Python dictionary. After this different operations can be applied, but at a different level: there are already many packages often provide deeper language specific functionality that should be leveraged. Example include UralicNLP \cite{uralicnlp_2019} for basic NLP analysis of Uralic languages, and murre for specific dialectal and historical text normalization or lemmatization scenarios \cite{537e4ad5fc054610a23474d91532bb07,agricolalemmatizer,334ff797181d4746a86028ca81ef4105}. The NLP for Latin also seems fairly developed, and available models could be applied \cite{thibault_clerice_2021_4661034}. 
%For larger languages general NLP tools such as spaCy are naturally immediately functional alternative.\footnote{https://spacy.io} 
We see as specific challenges in this the multilinguality and the continuous presence of words and expressions in different languages. %in other languages than Latin. 

%One task which we would see as closely connected to processing this type of texts is, however, transliteration. It is common that distinct researchers have published large amounts of materials in their own specific transcription systems, which makes the text entirely unaccessible for general NLP tools. 
%This has not been yet implemented for Castrén, primarily because his transcription is not phonemic in the sense of contemporary linguistic description, which means it cannot be easily converted. %However, we do publish an additional conversion pattern written by Jack Rueter that handles effectively similar transcription to orthography conversion for Komi materials from the 1960s.
%However, the sentence aligned data set is included to the publication exactly for the purpose of being able to test different transliteration and keyword detection pipelines in this context.

\section{Further usage}

The materials we have discussed have been created for two purposes: 1) openly licensed ground truth material for text recognition models, and 2) recognized, manually corrected, text for ethnographic and linguistic research. Text recognition models are at the moment line-based and the latter mainly relies on the subject knowledge of the researcher. Neither of these tasks necessarily demands further automatic processing of the materials, at least as long as the research is based on visual use of original versions and the recognized text is used as an aid and search tool to navigate in the document. 

However, we consider it still extremely important to be able to extract the text correctly from the files. In the current dataset both the line and layout element structure is indicated by order numbers, and simple concatenation of the lines thereby, in principle, yields the wanted order. However, there are cases where the situation is more complicated. The running order of the elements and lines may not be manually corrected, and it relies on individual conventions whether there is some way to mark whether the order has been verified manually. 

In our dataset we find that the running order of the text is generally correct at the page level, and especially so in document pages where layout is simple. This includes the printed Komi-Zyrian grammar and ethnographic notes. In the former table layout detection would need attention, and in the latter problems arise primarily from margin notes and comments between lines. Currently those are not easily placed to the correct locations in the text. %, mainly because we have not always indicated in the annotations that these elements belong to marginalia. 
In complexly layouted documents with several columns we also find a question of how to indicate the relationship across the columns, as one line is often translation or comment of the other. %This would demand some type of a linking mechanism between the line elements that is not currently available. 
This issue is seen on almost all pages of the Tundra Nenets epic narrative dataset. 

% There are many problems caused from inability to extract correct order and document structure. We cannot reconstruct the complete text, which forces later approaches to focus into word level analysis. At the same time even this may not be possible if the word continues with a hyphen into a next line. %: this is probably the minimal context where the next line has to be correctly indicated for further automatic processing to be possible. 

\section{Conclusion}

Our experiments show that the currently available tools to process 19th century Latin grammar materials in an endangered language can be almost flawlessly recognized with out-of-the-box text recognition models. With handwritten materials the publicly available models need to be customized, but the current accuracy may give at least some starting point if the language and time period match. %When the handwritten collection is large enough the training HTR system becomes fast an important component in working effectively with these materials.
The divergence of transcription systems and their complex relation to the contemporary orthographies is one challenge that needs to be separately addressed.

%We publish with the final version of this study also the Python scripts used to extract the line and layout elements, along the three distinct datasets in Zenodo, which will allow further experimentation and evaluation. 

To advance actual NLP applications, we also suggest that a sample from Castrén's materials would be published as a treebank or other annotated structure. Such multilingual collection may not fit larger projects such as Universal Dependencies, but similar conventions and file structures could easily be used. How this can be connected to proofread Ground Truth resources, commentaries and digital editions is another question, but there are few materials better for testing this than Castrén's data that is openly available and still acutely relevant for contemporary research. 

%\section{Experiments}

%\subsection{Language detection}

%\subsection{Lemmatization}

%\subsection{Part of speech tagging}

%\section*{Acknowledgments}

%The acknowledgments should go immediately before the references. Do not number the acknowledgments section.
%Do not include this section when submitting your paper for review.

\bibliography{anthology,acl2020}
\bibliographystyle{acl_natbib}

\end{document}